\pdfoutput=1

\documentclass[11pt]{article}

\usepackage[]{emnlp2021}

\usepackage{times}
\usepackage{amsmath}
\usepackage{amssymb}
\usepackage{graphicx}
\usepackage{caption}
\usepackage{subcaption}
\usepackage{xcolor}
\usepackage{float}
\usepackage{booktabs}
\usepackage{natbib}
\usepackage{latexsym}
\usepackage[textsize=scriptsize]{todonotes}
\usepackage{algorithmicx}
\usepackage{hyperref}
\usepackage[ruled,vlined]{algorithm2e}
\usepackage{xr}
\usepackage[T1]{fontenc}

\usepackage[utf8]{inputenc}

\usepackage{microtype}

\title{Can NLI Models Verify QA Systems' Predictions?}

\author{Jifan Chen\ \ \ \ \ \ \ \ \ \ \ Eunsol Choi\ \ \ \ \ \ \ \ \ \ \ Greg Durrett \\
  Department of Computer Science \\
  The University of Texas at Austin \\
  \texttt{\{jfchen, eunsol, gdurrett\}@cs.utexas.edu}}

\begin{document}
\maketitle
\begin{abstract}
To build robust question answering systems, we need the ability to verify whether answers to questions are truly correct, not just ``good enough'' in the context of imperfect QA datasets. We explore the use of natural language inference (NLI) as a way to achieve this goal, as NLI inherently requires the premise (document context) to contain all necessary information to support the hypothesis (proposed answer to the question). We leverage large pre-trained models and recent prior datasets to construct powerful question conversion and decontextualization modules, which can reformulate QA instances as premise-hypothesis pairs with very high reliability. Then, by combining standard NLI datasets with NLI examples automatically derived from QA training data, we can train NLI models to evaluate QA systems' proposed answers. We show that our approach improves the confidence estimation of a QA model across different domains. Careful manual analysis over the predictions of our NLI model shows that it can further identify cases where the QA model produces the right answer for the wrong reason, i.e., when the answer sentence does not address all aspects of the question.

\end{abstract}

\begin{figure}[t]
\centering
\includegraphics[width=0.48\textwidth]{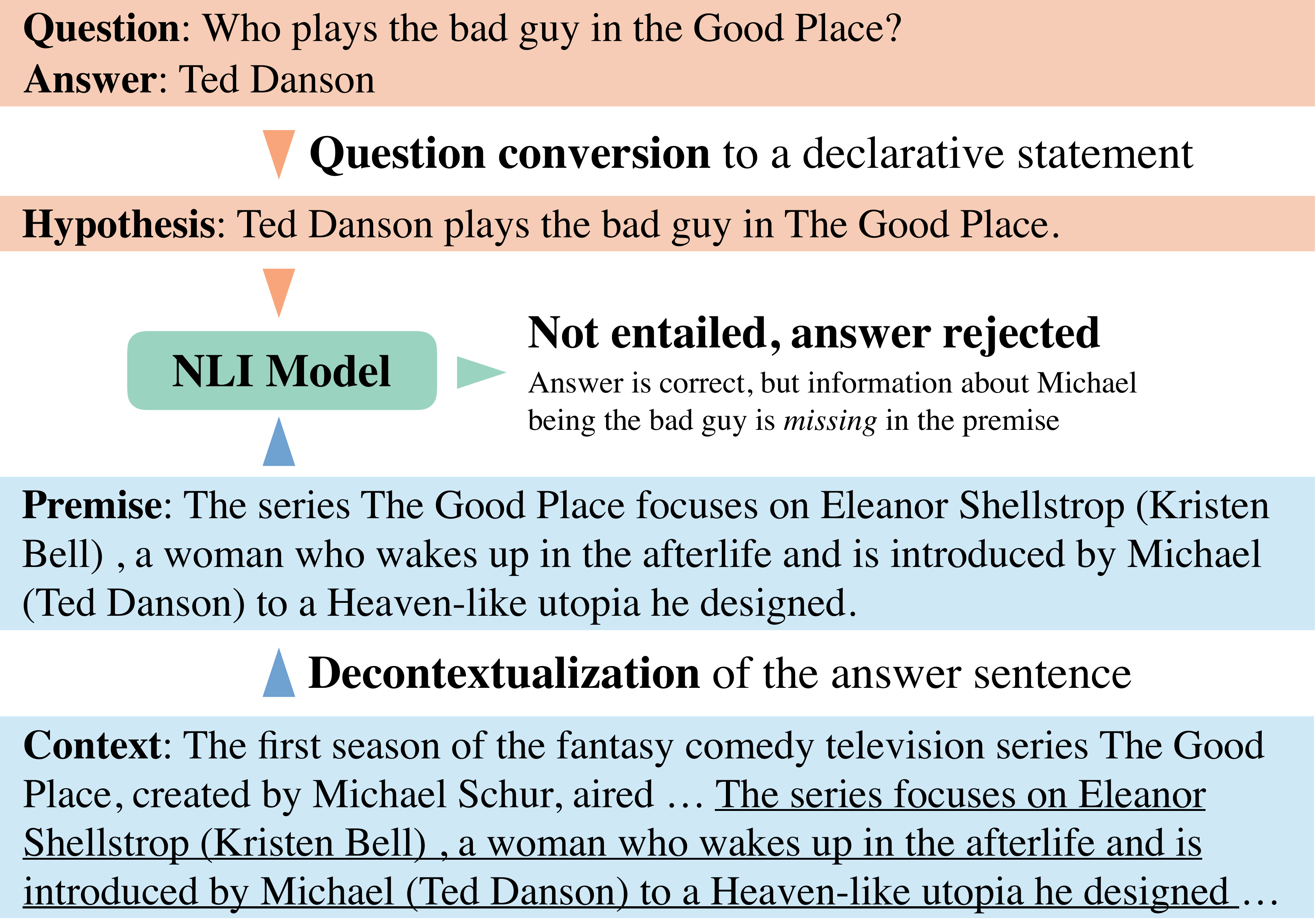}
\caption{An example from the Natural Questions dataset demonstrating how to convert a (question, context, answer) triplet to a (premise, hypothesis) pair. The underlined text denotes the sentence containing the answer \emph{Ted Danson}, which is then decontextualized by replacing \emph{The series} with \emph{The series The Good Place}. Although \emph{Ted Danson} is the right answer, an NLI model determines that the hypothesis is not entailed by the premise due to missing information.}
    \label{fig:processing_pipeline}
\end{figure}

\section{Introduction}

Recent question answering systems perform well on benchmark datasets~\cite{Seo2017BidirectionalAF,devlin2019bert,guu2020realm}, but these models often lack the ability to verify whether an answer is correct or not; they can correctly reject some unanswerable questions \cite{rajpurkar2018know,kwiatkowski2019natural,asai-choi-2021-challenges}, but are not always well-calibrated to spot spurious answers under distribution shifts~\cite{jia-liang-2017-adversarial,kamath-etal-2020-selective}. Natural language inference (NLI)~\cite{dagan2005pascal,bowman2015large} suggests one way to address this shortcoming: logical entailment provides a more rigorous notion for when a hypothesis statement is entailed by a premise statement. By viewing the answer sentence in context as the premise, paired with the question and its proposed answer as a hypothesis (see Figure~\ref{fig:processing_pipeline}), we can use NLI systems to verify that the answer proposed by a QA model satisfies the entailment criterion \cite{harabagiu-hickl-2006-methods,richardson-etal-2013-mctest}. 

Prior work has paved the way for this application of NLI. Pieces of our pipeline like converting a question to a declarative sentence \cite{wang-etal-2018-glue, demszky2018transforming} and reformulating an answer sentence to stand on its own \cite{choi2021decontextualization} have been explored. Moreover, an abundance of NLI datasets \cite{bowman2015large,williams2018broad} and related fact verification datasets~\cite{thorne2018fever} provide ample resources to train reliable models. We draw on these tools to enable NLI models to verify the answers from QA systems, and critically investigate the benefits and pitfalls of such a formulation.

Mapping QA to NLI enables us to exploit both NLI and QA datasets for answer verification, but as Figure~\ref{fig:processing_pipeline} shows, it relies on a pipeline for mapping a (question, answer, context) triplet to a (premise, hypothesis) NLI pair. We implement a strong pipeline here: we extract a concise yet sufficient premise through decontextualization~\cite{choi2021decontextualization}, which rewrites a single sentence from a document such that it can retain the semantics when presented alone without the document. We improve a prior question conversion model \cite{demszky2018transforming} with a stronger pre-trained seq2seq model, namely T5 \cite{raffel2020exploring}. Our experimental results show that both steps are critical for mapping QA to NLI. Furthermore, our error analysis shows that these two steps of the process are quite reliable and only account for a small fraction of the NLI verification model's errors.

Our evaluation focuses on two factors. First, can NLI models be used to improve calibration of QA models or boost their confidence in their decisions? Second, how does the entailment criterion of NLI, which is defined somewhat coarsely by crowd annotators \cite{williams2018broad}, transfer to QA? We train a QA model on Natural Questions~\cite[NQ]{kwiatkowski2019natural} and test whether using an NLI model helps it better generalize to four out-of-domain datasets from the MRQA shared task~\cite{fisch-etal-2019-mrqa}. We show that by using the question converter, the decontextualization model, and the automatically generated NLI pairs from QA datasets, \textbf{our NLI model improves the calibration over the base QA model across five different datasets.}\footnote{The converted NLI datasets, the question converter, the decontextualizer, and the NLI model are available at  ~\url{https://github.com/jifan-chen/QA-Verification-Via-NLI} } For example, in the selective QA setting~\cite{kamath-etal-2020-selective}, our approach improves the F1 score of the base QA model from 81.6 to 87.1 when giving answers on the 20\% of questions it is most confident about. Our pipeline further identifies the cases where there exists an information mismatch between the premise and the hypothesis. We find that existing QA datasets encourage models to return answers when the context does not actually contain sufficient information, suggesting that fully verifying the answers is a challenging endeavor.

\section{Using NLI as a QA Verifier}

\subsection{Background and Motivation}

Using entailment for QA is an old idea; our high-level approach resembles the approach discussed in \citet{harabagiu-hickl-2006-methods}. Yet, the execution of this idea differs substantially as we exploit modern neural systems and newly proposed annotated data for passage and question reformulation. \citet{richardson-etal-2013-mctest} explore a similar pipeline, but find that it works quite poorly, possibly due to the low performance of entailment systems at the time \cite{stern-dagan-2011-confidence}. We believe that a combination of recent advances in natural language generation \cite{demszky2018transforming,choi2021decontextualization} and strong models for NLI~\cite{liu2019roberta} equip us to re-evaluate this approach.

Moreover, the focus of other recent work in this space has been on transforming QA \emph{datasets} into NLI \emph{datasets}, which is a different end. \citet{demszky2018transforming} and \citet{mishra-etal-2021-looking} argue that QA datasets feature more diverse reasoning and can lead to stronger NLI models, particularly those better suited to strong contexts, but less attention has been paid to whether this agrees with classic definitions of entailment \cite{dagan2005pascal} or short-context NLI settings \cite{williams2018broad}.

Our work particularly aims to shed light on \textbf{information sufficiency} in question answering. Other work in this space has focused on validating answers to unanswerable questions \cite{rajpurkar2018know,kwiatkowski2019natural}, but such questions may be nonsensical in context; these efforts do not address whether all aspects of a question have been covered. Methods to handle adversarial SQuAD examples \cite{jia-liang-2017-adversarial} attempt to do this \cite{chen2021robust}, but these are again geared towards detecting specific kinds of mismatches between examples and contexts, like a changed modifier of a noun phrase. \citet{kamath-etal-2020-selective} frame their selective question answering techniques in terms of spotting out-of-domain questions that the model is likely to get wrong rather than more general confidence estimation. What is missing in these threads of literature is a formal criterion like entailment: when is an answer truly sufficient and when are we confident that it addresses the question?

\subsection{Our Approach}

Our pipeline consists of an answer candidate generator, a question converter, and a decontextualizer, which form the inputs to the final entailment model.

\paragraph{Answer Generation}
In this work, we focus our attention on extractive QA~\cite{hermann2015teaching,rajpurkar-etal-2016-squad}, for which we can get an answer candidate by running a pre-trained QA model.\footnote{Our approach could be adapted to multiple choice QA, in which case this step could be omitted.} We use the \texttt{Bert-joint} model proposed by~\newcite{alberti2019bert} for its simplicity and relatively high performance.

\paragraph{Question Conversion}
Given a question $q$ and an answer candidate $a$, our goal is to convert the $(q,a)$ pair to a declarative answer sentence $d$ which can be treated as the hypothesis in an NLI system~\cite{demszky2018transforming, khot2018scitail}. While rule-based approaches have long been employed for this purpose \cite{cucerzanagichtein2005}, the work of \citet{demszky2018transforming} showed a benefit from more sophisticated neural modeling of the distribution $P(d \mid q, a)$. We fine-tune a seq2seq model, T5-3B~\cite{raffel2020exploring}, using the $(a, q, d)$ pairs annotated by~\newcite{demszky2018transforming}.

While the conversion is trivial on many examples (e.g., replacing the wh-word with the answer and inverting the wh-movement), we see improvement on challenging examples like the following NQ question: \emph{the first vice president of India who became the president later was?} The rule-based system from~\citet{demszky2018transforming} just replaces \emph{who} with the answer \emph{Venkaiah Naidu}. Our neural model successfully appends the answer to end of the question and gets the correct hypothesis.

\paragraph{Decontextualization}
Ideally, the full context containing the answer candidate could be treated as the premise to make the entailment decision. But the full context often contains many irrelevant sentences and is much longer than the premises in single-sentence NLI datasets~\cite{williams2018broad, bowman2015large}. This length has several drawbacks. First, it makes transferring models from the existing datasets challenging. Second, performing inference over longer forms of text requires a multitude of additional reasoning skills like coreference resolution, event detection, and abduction \cite{mishra-etal-2021-looking}. Finally, the presence of extraneous information makes it harder to evaluate the entailment model's judgments for correctness; in the extreme, we might have to judge whether a fact about an entity is true based on its entire Wikipedia article, which is impractical.

We tackle this problem by \emph{decontextualizing} the sentence containing the answer from the full context to make it stand alone. Recent work~\cite{choi2021decontextualization} proposed a sentence decontextualization task in which a sentence together with its context are taken and the sentence is rewritten to be interpretable out of context if feasible, while preserving its meaning. This procedure can involve name completion (e.g., \emph{Stewart} $\rightarrow$ \emph{Kristen Stewart}), noun phrase/pronoun swap, bridging anaphora resolution, and more.

More formally, given a sentence $S_a$ containing the answer and its corresponding context $C$, decontextualization learns a model $P(S_d \mid S_a, C)$, where $S_d$ is the decontextualized form of $S_a$. We train a decontextualizer by fine-tuning the T5-3B model to decode $S_d$ from a concatenation of $(S_a, C)$ pair, following the original work. More details about the models we discuss here can be found in Appendix~\ref{section:appendix:model_details}.

\section{Experimental Settings}
\label{sec:experimental_setting}
Our experiments seek to validate the utility of NLI for verifying answers \textbf{primarily under distribution shifts}, following recent work on selective question answering \cite{kamath-etal-2020-selective}. We transfer an NQ-trained QA model to a range of datasets and evaluate whether NLI improves answer confidence.

\paragraph{Datasets} We use five English-language span-extractive QA datasets: Natural Questions \cite[NQ]{kwiatkowski2019natural}, TriviaQA~\cite{JoshiTriviaQA2017}, BioASQ~\cite{bioasq}, Adversarial SQuAD \cite[SQuAD-adv]{jia-liang-2017-adversarial}, and SQuAD 2.0~\cite{rajpurkar2018know}. For TriviaQA and BioASQ, we use processed versions from MRQA~\cite{fisch-etal-2019-mrqa}. These datasets cover a wide range of domains including biology (BioASQ), trivia questions (TriviaQA), real user questions (NQ), and human-synthetic challenging sets (SQuAD2.0 and SQuAD-adv). 
For NQ, we filter out the examples in which the questions are narrative statements rather than questions by the rule-based system proposed by~\newcite{demszky2018transforming}. We also exclude the examples based on tables because they are not compatible with the task formulation of NLI.\footnote{After filtering, we have 191,022/4,855 examples for the training and development sets respectively. For comparison, the original NQ contains 307,373/7,842 examples for training and development.}

\paragraph{Base QA Model} We train our base QA model~\cite{alberti2019bert} with the NQ dataset. To study robustness across different datasets, we fix the base QA model and investigate its capacity to transfer. We chose NQ for its high quality and the diverse topics it covers. 

\paragraph{Base NLI Model} We use the RoBERTa-based NLI model trained using Multi-Genre Natural Language Inference~\cite[MNLI]{williams2018broad} from AllenNLP~\cite{gardner-etal-2018-allennlp} for its broad coverage and high accuracy. 

\paragraph{QA-enhanced NLI Model}
As there might exist different reasoning patterns in the QA datasets which are not covered by the MNLI model~\cite{mishra-etal-2021-looking}, we study whether NLI pairs \emph{generated from QA datasets} can be used jointly with the MNLI data to improve the performance of an NLI model. To do so, we run the QA instances in the NQ training set through our QA-to-NLI conversion pipeline, resulting in a dataset we call \textbf{NQ-NLI}, containing (premise, hypothesis) pairs from NQ with binary labels. As answer candidates, we use the predictions of the base QA model. If the predicted answer is correct, we label the (premise, hypothesis) as positive (entailed), otherwise negative (not entailed). To combine NQ-NLI with MNLI, we treat the examples in MNLI labeled with ``entailment'' as positive and the others as negative. We take the same number of examples as of NQ-NLI from MNLI and shuffle them to get a mixed dataset which we call \textbf{NQ-NLI+MNLI}. We use these dataset names to indicate NLI models trained on these datasets.

Some basic statistics for each dataset after processing with our pipeline are shown in Appendix~\ref{section:appendix:dataset_statistics}. 

\begin{figure}[t]
\centering
\includegraphics[width=0.48\textwidth]{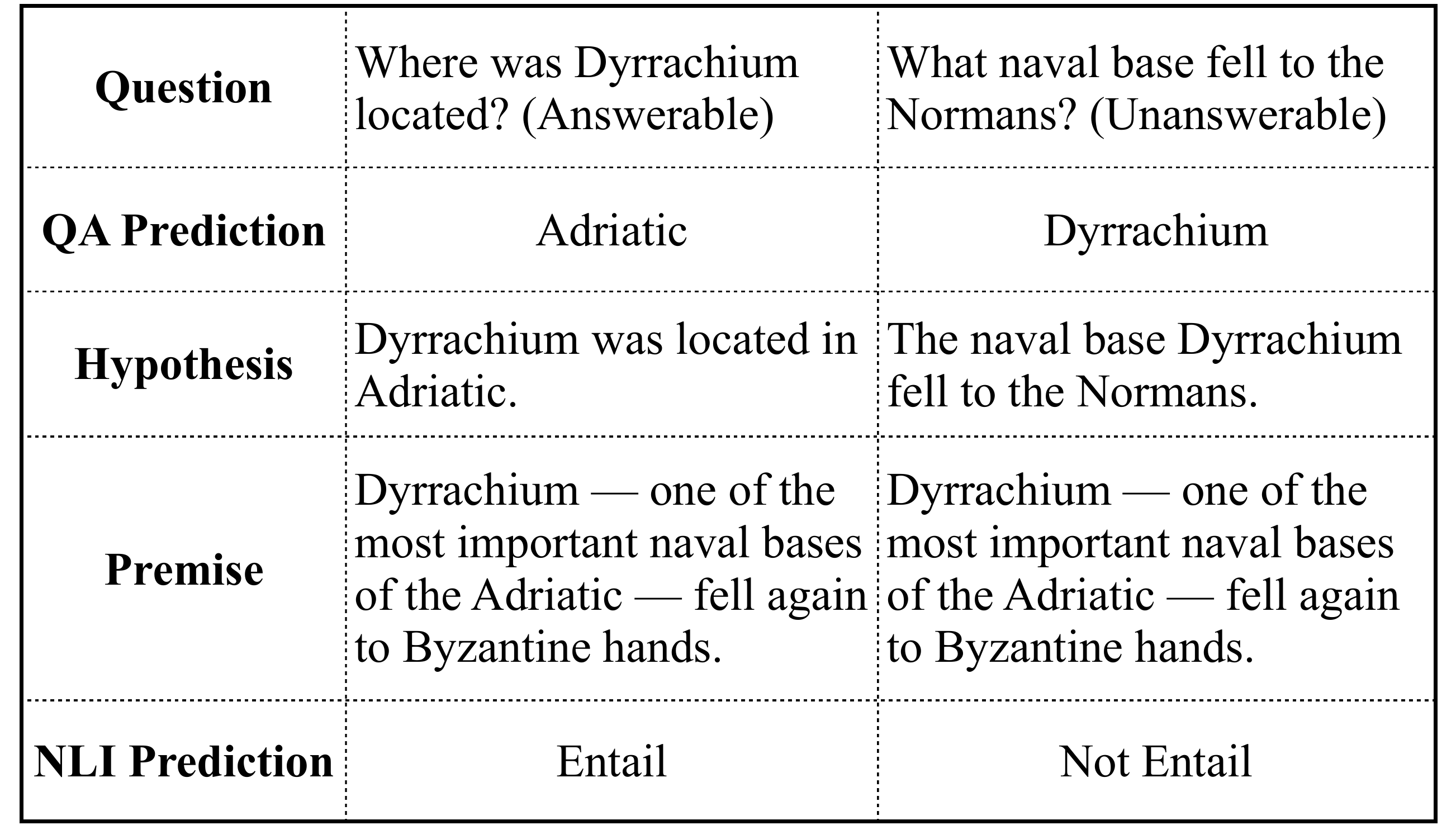}
\caption{Two examples from SQuAD2.0. The MNLI model successfully accepts the correct answer for the answerable question (left) and rejects a candidate answer for the unanswerable one (right).}
    \label{fig:squad2.0_examples}
\end{figure}

\section{Improving QA Calibration with NLI}
In this section, we explore to what extent either off-the-shelf or QA-augmented NLI models work as verifiers across a range of QA datasets.

\subsection{Rejecting Unanswerable Questions}

We start by testing how well a pre-trained MNLI model, with an accuracy of 90.2\% on held-out MNLI examples, can identify unanswerable questions in SQuAD2.0. We run our pre-trained QA model on the unanswerable questions to produce answer candidates, then convert them to the NLI pairs through our pipeline, including question conversion and decontextualization. We run the entailment model trained on MNLI to see how frequently it is able to reject the answer by predicting either ``neutral'' or ``contradiction''. For questions with annotated answers, we also generate the NLI pairs with the gold answer and see if the entailment model trained on MNLI can accept the answer.

The MNLI model successfully rejects \textbf{78.5\%} of the unanswerable examples and accepts \textbf{82.5\%} of the answerable examples. Two examples taken from SQuAD2.0 are shown in Figure~\ref{fig:squad2.0_examples}. We can see the MNLI model is quite sensitive to the information mismatch between the hypothesis and the premise. In the case where there is no information about \emph{Normans} in the premise, it rejects the answer. Without seeing any data from SQuAD2.0, MNLI can already act as a strong verifier in the unanswerable setting where it is hard for a QA model to generalize~\cite{rajpurkar2018know}.

\subsection{Calibration}
To analyze the effectiveness of the NLI models in a more systematic way, we test whether they can improve calibration of QA models or improve model performance in a ``selective'' QA setting~\cite{kamath-etal-2020-selective}. That is, \textbf{if our model can choose to answer only the $k$ percentage of examples it is most confident about (the coverage), what F1 can it achieve?} We first rank the examples by the confidence score of a model; for our base QA models, this score is the posterior probability of the answer span, and for our NLI-augmented models, it is the posterior probability associated with the ``entailment'' class. We then compute F1 scores at different coverage values.

\begin{figure}[t]
\centering
\includegraphics[width=0.45\textwidth]{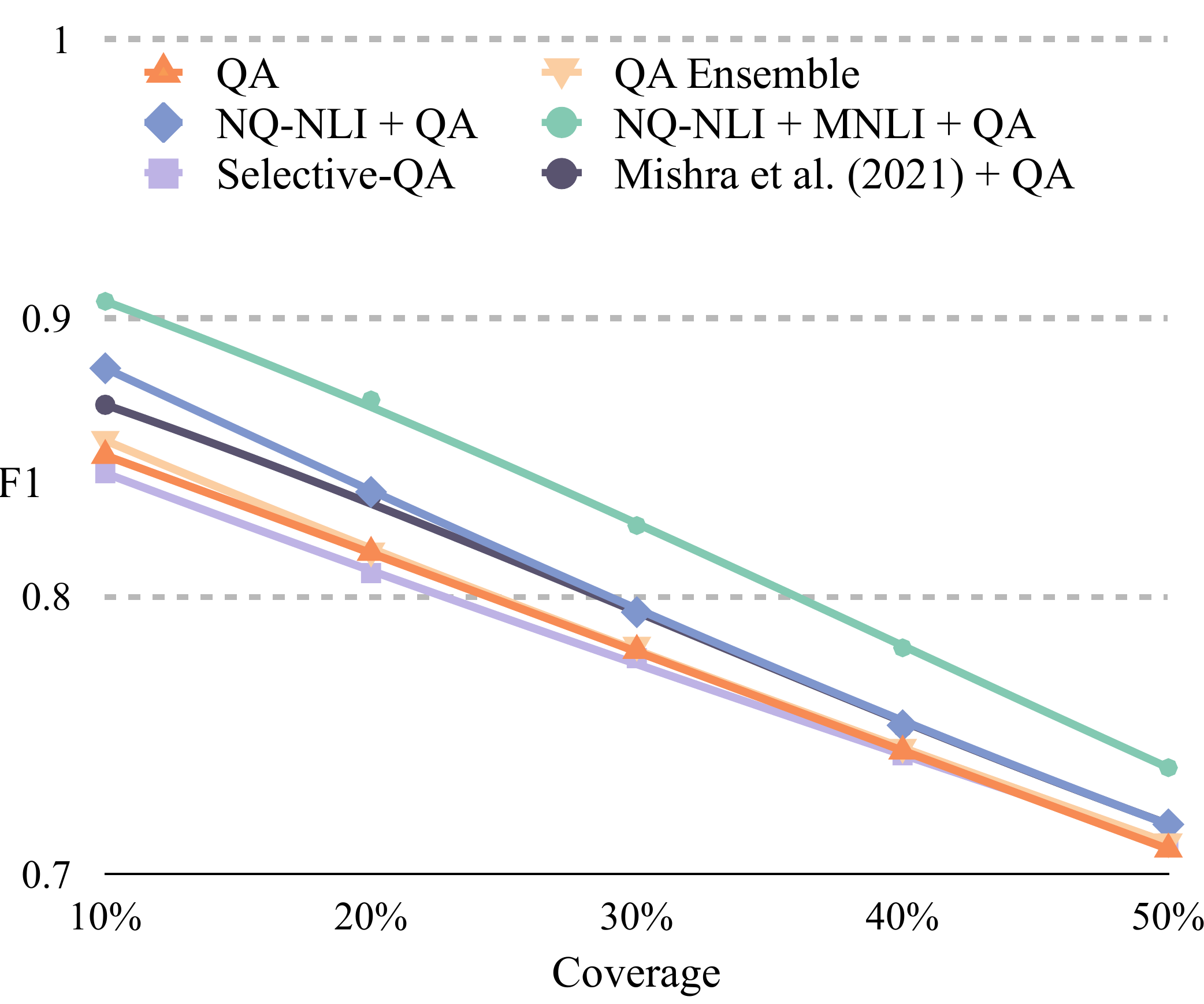}
\caption{Average calibration performance of our models \emph{combining the posterior} from the NQ-NLI and the QA models over five datasets. The x-axis denotes the top $k\%$ of examples the model is answering, ranked by the confidence score. The y-axis denotes the F1 score.}
    \label{fig:exp_qa_nli_fine_tuning_plus_qa_avg}
\end{figure}

\begin{figure}[t]
\centering
\includegraphics[width=0.45\textwidth]{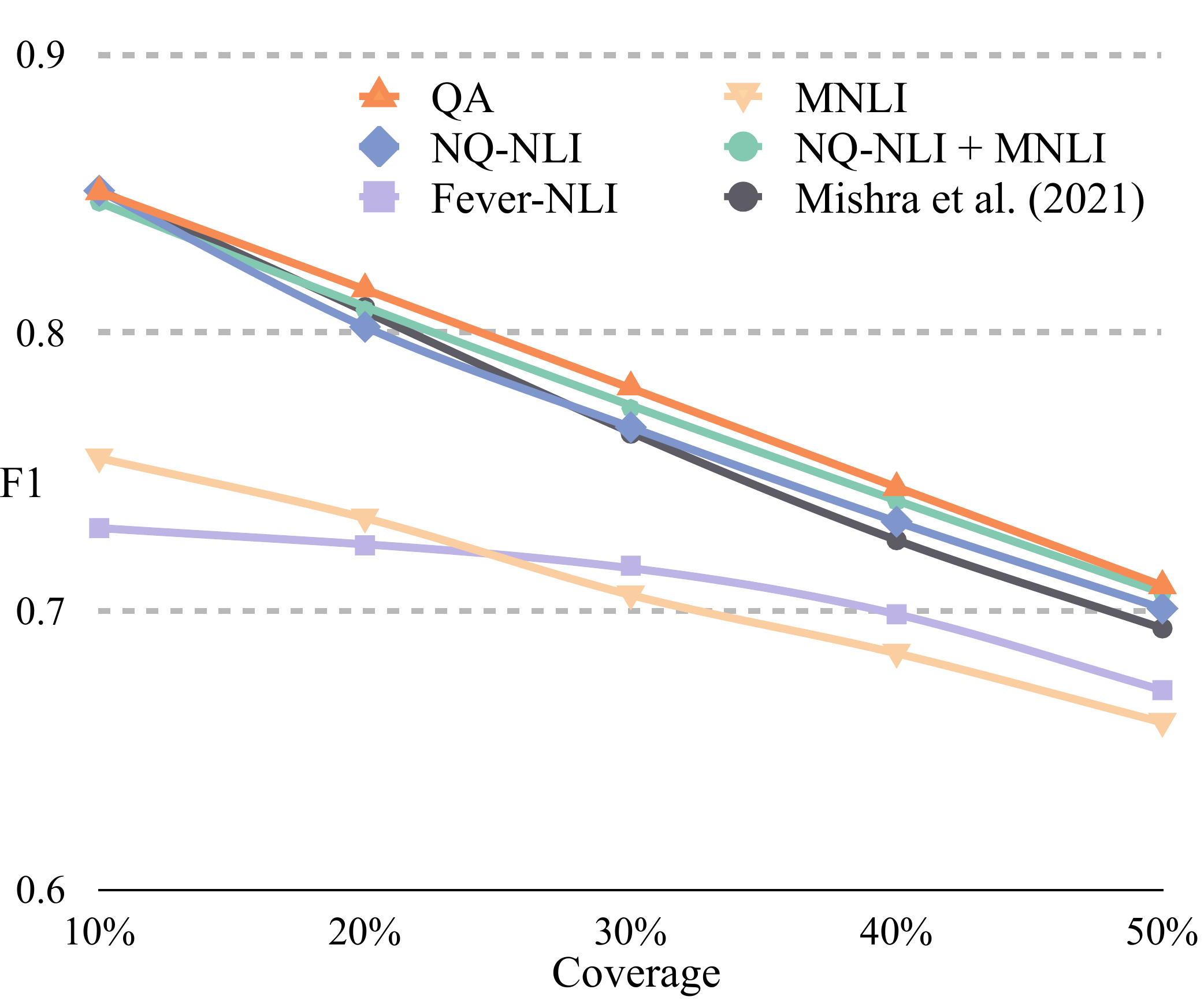}
\caption{Average calibration performance of our NLI models \emph{alone} (not including QA posteriors) trained on NQ-NLI over five datasets. The x-axis denotes the top $k\%$ of examples the model is answering, ranked by the confidence score. The y-axis denotes the F1 score.}
    \label{fig:exp_qa_nli_fine_tuning_avg}
\end{figure}

\subsubsection{Comparison Systems}

\paragraph{NLI model variants} We train separate NLI models with MNLI, NQ-NLI, NQ-NLI+MNLI introduced in Section~\ref{sec:experimental_setting}, as well as with the NLI version of the FEVER~\cite{thorne2018fever} dataset, which is retrieved by~\newcite{nie2019combining}. As suggested by~\newcite{mishra-etal-2021-looking}, an NLI model could benefit from training with premises of different length; therefore, we train an NLI model without the decontextualization phase of our pipeline on the combined data from both NQ-NLI and MNLI. We call this model \textbf{Mishra et al. (2021)} since it follows their procedure. All of the models are initialized using RoBERTa-large~\cite{liu2019roberta} and trained using the same configurations.

\paragraph{NLI+QA} We explore combining complementary strengths of the NLI posteriors and the base QA posteriors. We take the posterior probability of the two models as features and learn a binary classifier $y= \textrm{logistic}(w_1p_{\textrm{QA}} + w_2p_{\textrm{NLI}})$ as the combined entailment model and tune the model on 100 held-out NQ examples. \textbf{+QA} denotes this combination with any of our NLI models.

\paragraph{QA-Ensemble} To compare with \textbf{NLI+QA}, we train another identical QA model, \texttt{Bert-joint}, using the same configurations and ensemble the two QA models using the same way as \textbf{NLI+QA}.

\paragraph{Selective QA}
~\newcite{kamath-etal-2020-selective} train a calibrator to make models better able to selectively answer questions in new domains. The calibrator is a binary classifier with seven features: passage length, the length of the predicted answer, and the top five softmax probabilities output by the QA model. We use the same configuration as~\cite{kamath-etal-2020-selective} and train the calibrator on the same data as our NQ-NLI model.

\begin{figure}[t]
\centering
\includegraphics[width=0.48\textwidth]{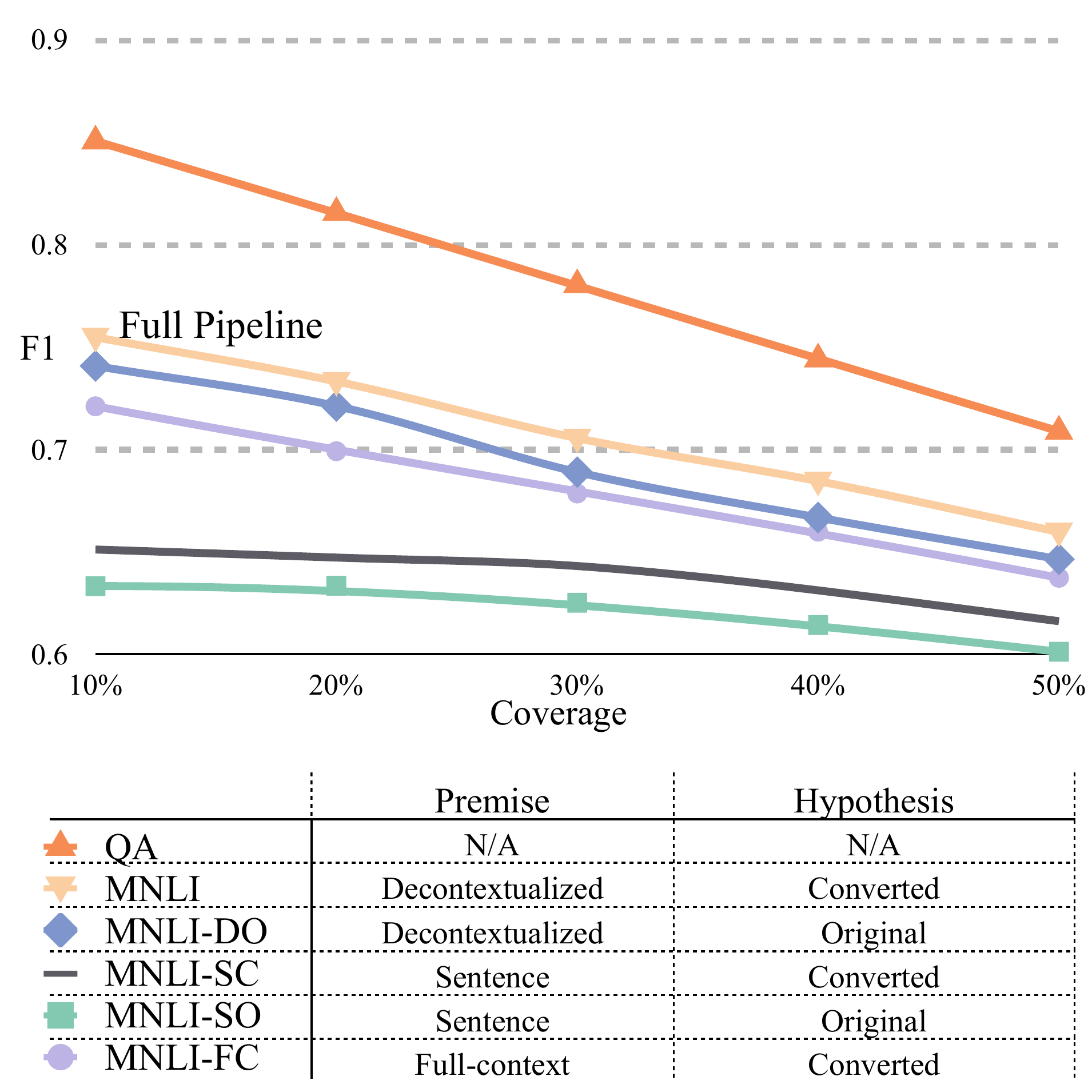}
\caption{Average calibration performance of the MNLI model on five QA datasets. Converted vs.~original denotes using the converted question or the original question concatenated with the answer as the hypothesis. Sentence vs.~decontextualized vs.~full-context denotes using the sentence containing the answer, its decontextualized form, or the full context as the premise. }
    \label{fig:exp_MNLI_zero_shot_avg}
\end{figure}

\begin{figure*}[t]
\centering
\includegraphics[width=1.0\textwidth]{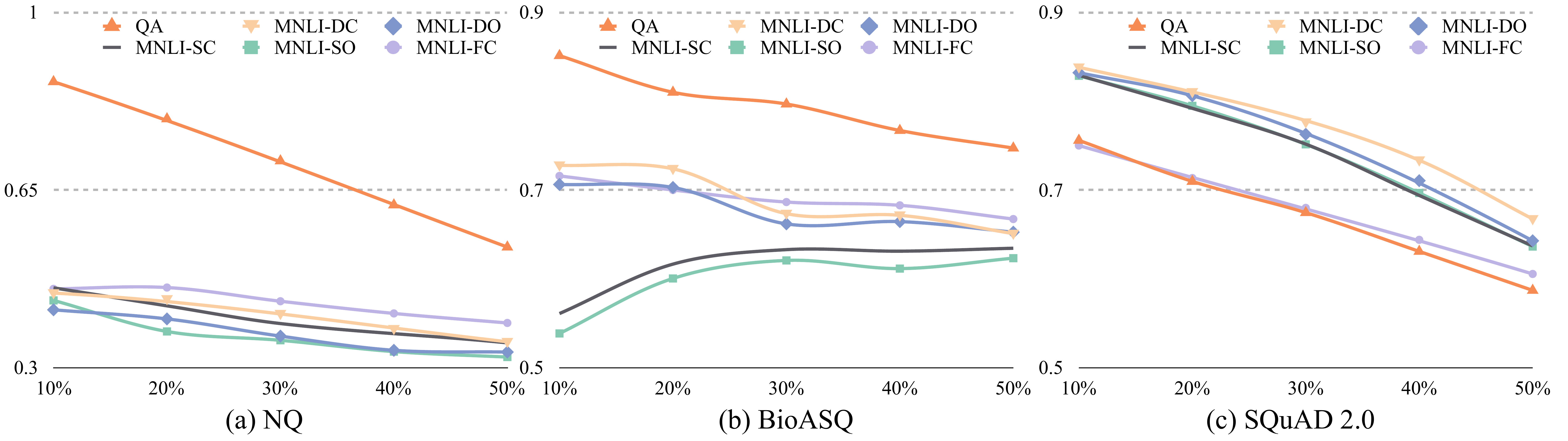}
\caption{Calibration performance of the MNLI model on three out of five QA datasets we used. Here, we omit TriviaQA and SQuAD-adv since they exhibit similar behavior as BioASQ and SQuAD2.0, respectively. The legends share the same semantics as Figure~\ref{fig:exp_MNLI_zero_shot_avg}. The x-axis denotes coverage and the y-axis denotes the F1 score. }
    \label{fig:exp_MNLI_zero_shot}
\end{figure*}

\subsubsection{Results and Analysis}
Figure~\ref{fig:exp_qa_nli_fine_tuning_plus_qa_avg} shows the macro-averaged results over the five QA datasets. Please refer to Appendix~\ref{section:appendix:performance-breakdown} for per dataset breakdown.

Our \textbf{NQ-NLI+QA} system, which combines the QA models' posteriors with an NQ-NLI-trained system, already shows improvement over using the base QA posteriors. Surprisingly, additionally training the NLI model on MNLI (\textbf{NQ-NLI+MNLI+QA}) gives \emph{even stronger} results. The NLI models appear to be complementary to the QA model, improving performance even on out-of-domain data. We also see that our our \textbf{NQ-NLI+MNLI+QA} outperforms \textbf{Mishra et al.~(2021)+QA} by a large margin. By inspecting the performance breakdown in Appendix~\ref{section:appendix:performance-breakdown}, we see the gap is mainly on SQuAD2.0 and SQuAD-adv. This is because these datasets often introduce subtle mismatches by slight modification of the question or context; even if the NLI model is able to overcome other biases, these are challenging contrastive examples from the standpoint of the NLI model. This observation also indicates that to better utilize the complementary strength of MNLI, the proposed decontextualization phase in our pipeline is quite important.  

\textbf{Selective QA} shows similar performance to using the posterior from QA model, which is the most important feature for the calibrator.

Combining NLI model with the base QA models' posteriors is necessary for this strong performance. Figure~\ref{fig:exp_qa_nli_fine_tuning_avg} shows the low performance achieved by the NLI models alone, indicating that \textbf{NLI models trained exclusively on NLI dataset (FEVER-NLI, MNLI) cannot be used by themselves as effective verifiers for QA}. 
This also indicates a possible domain or task mismatch between FEVER, MNLI, and the other QA datasets.

\textbf{NQ-NLI helps bridge the gap between the QA datasets and MNLI.} In Figure~\ref{fig:exp_qa_nli_fine_tuning_avg}, both NQ-NLI and NQ-NLI+MNLI achieve similar performance to the original QA model. We also find that training using both NQ-NLI and MNLI achieves slightly better performance than training using NQ-NLI alone. This suggests that we are not simply training a QA model of a different form by using the NQ-NLI data; rather, the NQ-NLI pairs are compatible with the MNLI pairs, and the MNLI examples are useful for the model.


\section{Effectiveness of the Proposed Pipeline}
We present an ablation study on our pipeline to see how each component contributes to the final performance. For simplicity, we use the off-the-shelf {MNLI} model since it does not involve training using the data generated through the pipeline. Figure~\ref{fig:exp_MNLI_zero_shot_avg} shows the average results across five datasets and Figure~\ref{fig:exp_MNLI_zero_shot} presents individual performance on three datasets.

We see that \textbf{both the question converter and the decontextualizer contribute to the performance of the MNLI model.} In both figures, removing either module harms the performance for all datasets. On NQ and BioASQ, using the full context is better than the decontextualized sentence, which hints that there are cases where the full context provides necessary information. We have a more comprehensive analysis in Section~\ref{sec:error_analysis_entailment}.

Moreover, we see that {MNLI outperforms the base QA posteriors on SQuAD2.0 and SQuAD-adv.} Figure~\ref{fig:exp_MNLI_zero_shot}(a) also shows that the largest gap between the QA and NLI model is on NQ, which is unsurprising since the QA model is trained on NQ. These results show how the improvement in the last section is achieved: the complementary strengths of  MNLI and NQ datasets lead to the best overall performance.

\section{Understanding the Behavior of NQ-NLI}
We perform manual analysis on 300 examples drawn from NQ, TriviaQA, and SQuAD2.0 datasets where \textbf{NQ-NLI+MNLI} model produced an error. We classify errors into one of 7 classes, described in Section~\ref{sec:errors_pipeline} and \ref{sec:error_analysis_entailment}. All of the authors of this paper conducted the annotation. The annotations agree with a Fleiss' kappa value of 0.78, with disagreements usually being between closely related categories among our 7 error classes, e.g., annotation error vs. span shifting, wrong context vs. insufficient context, as we will see later. The breakdown of the errors in each dataset is shown in Table~\ref{tab:error_type_breakdown}. 

\subsection{Errors from the Pipeline}
\label{sec:errors_pipeline}

\begin{figure}
\centering

\fbox{
\begin{minipage}[c]{0.47\textwidth}
\centering
    \includegraphics[width=3.0in]{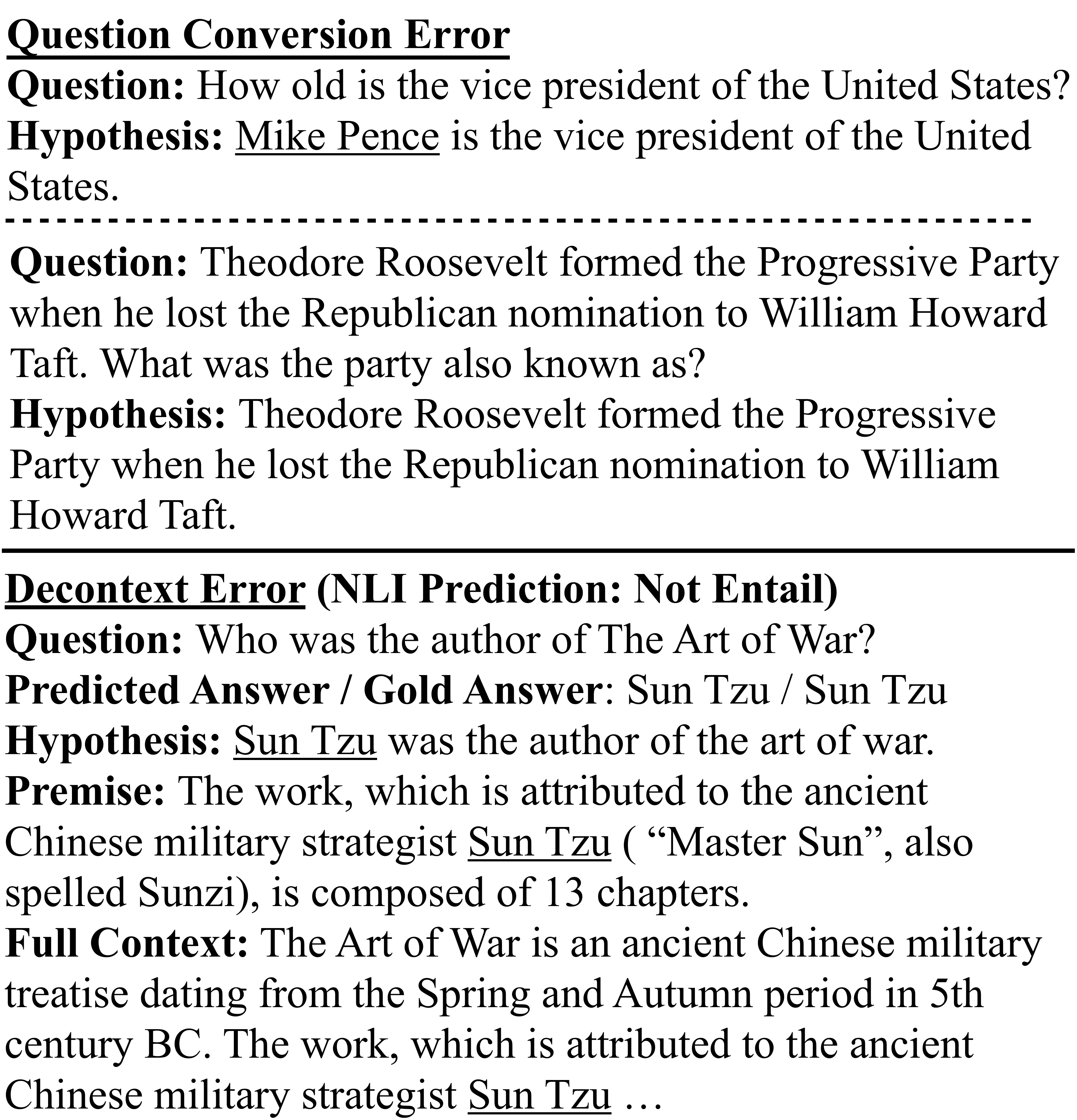}
\end{minipage}}
\caption{Pipeline error examples from the NQ development set: the underlined text span denotes the answer predicted by the QA model.}
    \label{fig:error_analysis_pipeline}
\end{figure}

We see that across the three different datasets, the number of errors attributed to our pipeline approach is below 10\%. This demonstrates that the question converter and the decontextualization model are quite effective to convert a (question, answer, context) triplet to a (premise, hypothesis) NLI pair. For the question converter, errors mainly happen in two scenarios as shown in Figure~\ref{fig:error_analysis_pipeline}. (1) The question converter gives an answer of the wrong type to a question. For example, the question asks ``\emph{How old...}'', but the answer returned is ``\emph{Mike Pence}'' which does not fit the question. The question converter puts \emph{Mike Pence} back into the question and yields an unrelated statement. Adding a presupposition checking stage to the question converter could further improve its performance~\cite{kim-etal-2021-linguist}. (2) The question is long and syntactically complex; the question converter just copies a long question without answer replacement.

For the decontextualization model, errors usually happen when the model fails to recall one of the required modifications. As shown in the example in Figure~\ref{fig:error_analysis_pipeline}, the model fails to replace \emph{The work} with its full entity name \emph{The Art of War}.
 
\subsection{Errors from the NLI Model}
\label{sec:error_analysis_entailment}
Most of the errors are attributed to the entailment model. We investigate these cases closely and ask ourselves \emph{if these really are errors}. We categorize them into the following categories.

\paragraph{Entailment} These errors are truly mistakes by the entailment model: in our view, the pair of sentences should exhibit a different relationship than what was predicted.

\begin{figure*}[t]
\centering
\includegraphics[width=1.0\textwidth]{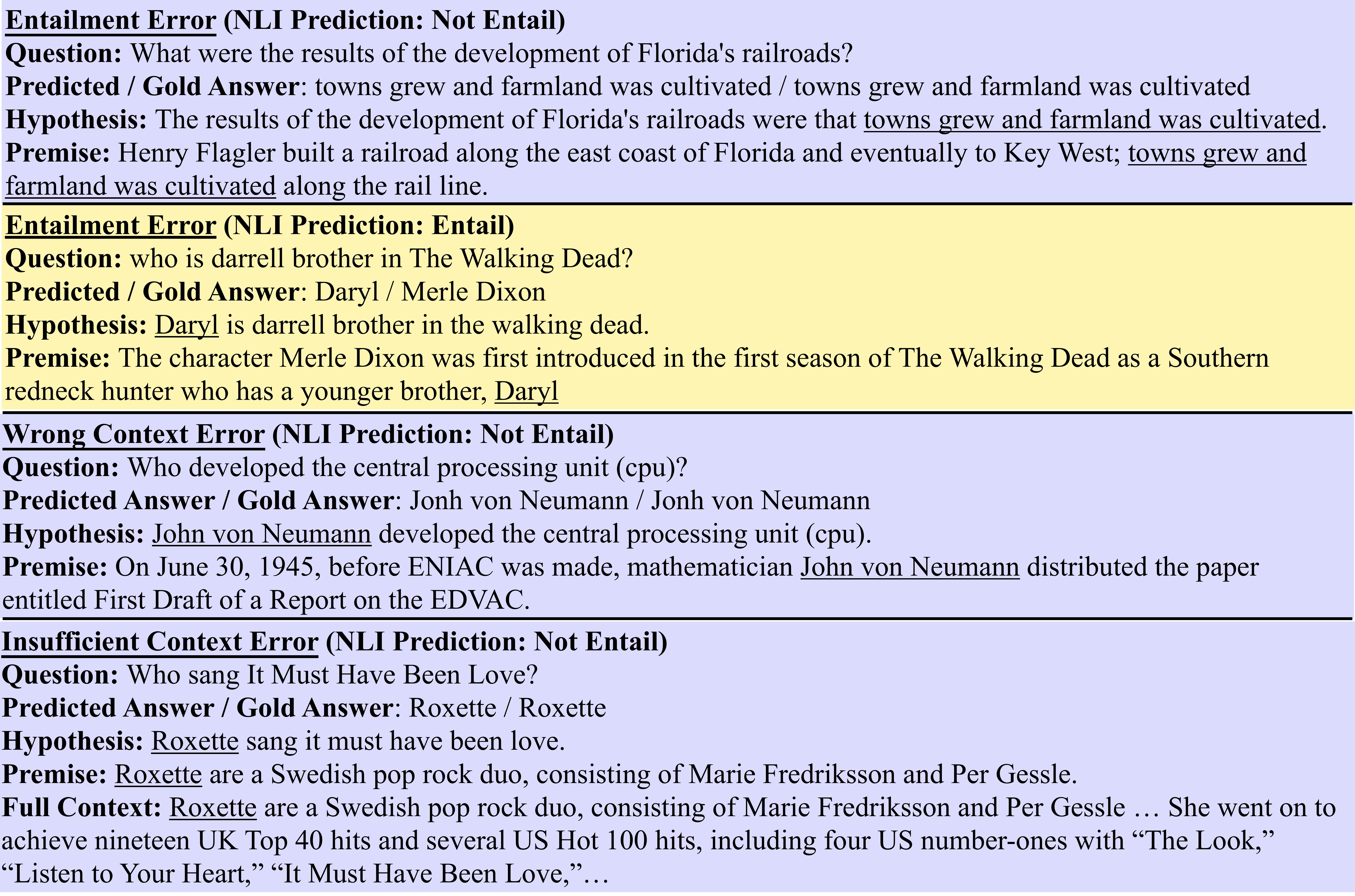}
\caption{Examples taken from the development sets of NQ and TriviaQA, grouped by different types of errors the entailment model makes. The underlined text span denotes the answer predicted by the QA model. The yellow box denotes a false positive example and the purple box denotes false negative examples.}
    \label{fig:error_analysis_NLI}
\end{figure*}

\paragraph{Wrong Context} The QA model gets the right answer for the wrong reason. The example in Figure~\ref{fig:error_analysis_NLI} shows that \emph{John Von Neumann} is the annotated answer but it is not entailed by the premise because no information about \emph{CPU} is provided. Although the answer is correct, {we argue it is better for the model to reject this case.} This again demonstrates one of the key advantages of using an NLI model as a verifier for QA models: it can identify cases of information mismatch like this where the model didn't retrieve suitable context to show to the user of the QA system.

\paragraph{Insufficient Context (out of scope for decontextualization)} The premise lacks essential information that could be found in the full context, typically later in the context. In Figure~\ref{fig:error_analysis_NLI}, the answer \emph{Roxette} is in the first sentence. However, we do not know that she wrote the song \emph{It Must Have Been Love} until we go further in the context. The need to add future information is beyond the scope of the decontextualization~\cite{choi2021decontextualization}.

\paragraph{Span Shifting} The predicted answer of the QA model overlaps with the gold answer and it is acceptable as a correct answer. For example, a question asks \emph{What Missouri town calls itself the Live Music Show Capital?} Both \emph{Branson} and \emph{Branson, Missouri} can be accepted as the right answer.  
 
\paragraph{Annotation Error} Introduced by the incomplete or wrong annotations -- some acceptable answers are missing or the annotated answer is wrong.

\begin{table}[t]
\footnotesize
\centering
\renewcommand{\tabcolsep}{1.7mm}
\begin{tabular}{ l | c c c }
\toprule
 & NQ &  TQA & SQuAD2.0 \\
\midrule
Question Conversion & \colorbox{yellow!60}{\phantom{0}3} \colorbox{blue!20}{\phantom{0}0} & \colorbox{yellow!60}{\phantom{0}0} \colorbox{blue!20}{\phantom{0}2} & \colorbox{yellow!60}{\phantom{0}2} \colorbox{blue!20}{\phantom{0}0} \\
Decontext & \colorbox{yellow!60}{\phantom{0}0} \colorbox{blue!20}{\phantom{0}4} & \colorbox{yellow!60}{\phantom{0}0} \colorbox{blue!20}{\phantom{0}0} & \colorbox{yellow!60}{\phantom{0}0} \colorbox{blue!20}{\phantom{0}7} \\
\midrule
Entailment & \colorbox{yellow!60}{12} \colorbox{blue!20}{39} & \colorbox{yellow!60}{\phantom{0}2} \colorbox{blue!20}{14} & \colorbox{yellow!60}{12} \colorbox{blue!20}{56}  \\
Wrong Context & \colorbox{yellow!60}{\phantom{0}0} \colorbox{blue!20}{23} & \colorbox{yellow!60}{\phantom{0}0} \colorbox{blue!20}{42} & \colorbox{yellow!60}{\phantom{0}0} \colorbox{blue!20}{\phantom{0}2}  \\
Insufficient Context &  \colorbox{yellow!60}{\phantom{0}0} \colorbox{blue!20}{11} & \colorbox{yellow!60}{\phantom{0}0} \colorbox{blue!20}{16} & \colorbox{yellow!60}{\phantom{0}0} \colorbox{blue!20}{\phantom{0}4} \\
Span Shifting & \colorbox{yellow!60}{\phantom{0}3} \colorbox{blue!20}{\phantom{0}0} & \colorbox{yellow!60}{13} \colorbox{blue!20}{\phantom{0}0} & \colorbox{yellow!60}{\phantom{0}7} \colorbox{blue!20}{\phantom{0}0}  \\
Annotation & \colorbox{yellow!60}{\phantom{0}5} \colorbox{blue!20}{\phantom{0}0} & \colorbox{yellow!60}{11} \colorbox{blue!20}{\phantom{0}0} & \colorbox{yellow!60}{10} \colorbox{blue!20}{\phantom{0}0} \\
\midrule
Total & \colorbox{yellow!60}{23} \colorbox{blue!20}{77} & \colorbox{yellow!60}{26} \colorbox{blue!20}{74} & \colorbox{yellow!60}{31} \colorbox{blue!20}{69} \\
\bottomrule
\end{tabular}
\caption{Error breakdown of our \textbf{NQ-NLI+MNLI} verifier on NQ, TQA (TriviaQA), and SQuAD2.0. Here, yellow and purple denote the false positive and false negative counts respectively. False positive: NLI predicts entailment while the answer predicted is wrong. False negative: NLI predicts non-entailment while the answer predicted is right.}

\label{tab:error_type_breakdown}
\end{table}

\noindent
From Table~\ref{tab:error_type_breakdown}, we see that ``wrong context'' cases consist of 25\% and 40\% of the errors for NQ and TriviaQA, respectively, while they rarely happen on SQuAD2.0. This is because the supporting snippets for NQ and TriviaQA are retrieved from Wikipedia and web documents, so the information contained may not be sufficient to support the question. For SQuAD2.0, the supporting document is given to the annotators, so no such errors happen.

This observation indicates that the NLI model can be particularly useful in the open-domain setting where it can reject answers that are not well supported. In particular, we believe that this raises a question about answers in TriviaQA. The supporting evidence for the answer is often \textbf{insufficient} to validate all aspects of the question. \textbf{What should a QA model do in this case: make an educated guess based on partial evidence, or reject the answer outright?} This choice is application-specific, but our approach can help system designers make these decisions explicit.

Around 10\% to 15\% of errors happens due to insufficient context. Such errors could be potentially fixed in future work by learning a question-conditioned decontextualizer which aims to gather all information related to the question.

\section{Related Work}

\paragraph{NLI for Downstream Tasks} ~\newcite{welleck2019dialogue} proposed a dialogue-based NLI dataset and the NLI model trained over it improved the consistency of a dialogue system; ~\newcite{pasunuru2017towards, li2018ensure, falke2019ranking} used NLI models to detect factual errors in abstractive summaries. For question answering, ~\newcite{harabagiu-hickl-2006-methods} showed that textual entailment can be used to enhance the accuracy of the open-domain QA systems; ~\newcite{trivedi2019repurposing} used a pretrained NLI model to select relevant sentences for multi-hop question answering; ~\newcite{yin-etal-2020-universal} tested whether NLI models generalize to QA setting in a few-shot learning scenario.

Our work is most relevant to ~\newcite{mishra-etal-2021-looking}; they also learn an NLI model using examples generated from QA datasets. Our work differs from theirs in a few chief ways. First, we improve the conversion pipeline significantly with decontextualization and a better question converter. Second, we use this framework to improve QA performance by using NLI as a verifier, which is only possible because the decontextualization allows us to focus on a single sentence. We also study whether the converted dataset is compatible with other off-the-shelf NLI datasets. By contrast, ~\newcite{mishra-etal-2021-looking} use their converted NLI dataset to aid other tasks such as fact-checking. Finally, the contrast we establish here allows us to conduct a thorough human analysis over the converted NLI data and show how the task specifications of NLI and QA are different (Section~\ref{sec:error_analysis_entailment}). 

\paragraph{Robust Question Answering}
Modern QA systems often give incorrect answers in challenging settings that require generalization~\cite{rajpurkar2018know, chen2019understanding, wallace-etal-2019-universal, gardner-etal-2020-evaluating, kaushik2019learning}. Models focusing on robustness and generalizability have been proposed in recent years: ~\newcite{wang2018robust,khashabi2020more,liu2020robust} use perturbation based methods and adversarial training; ~\newcite{lewis2018generative} propose generative QA to prevent the model from overfitting to simple patterns; ~\newcite{yeh2019qainfomax, zhou2020robust} use advanced regularizers; ~\newcite{clark2019don} debias the training set through ensemble-based training; and ~\newcite{chen2021robust} incorporate an explicit graph alignment procedure.

Another line of work to make models more robust is by introducing answer verification~\cite{hu2019read+,kamath-etal-2020-selective,wang-etal-2020-answer-better,zhang-etal-2021-knowing} as a final step for question answering models. Our work is in the same vein, but has certain advantages from using an NLI model. First, the answer verification process is more explicit so that one is able to spot where the error emerges. Second, we can incorporate NLI datasets from other domains into the training of our verifier, reducing reliance on in-domain labeled QA data.

\section{Conclusion}

This work presents a strong pipeline for converting QA examples into NLI examples, with the intent of verifying the answer with NLI predictions. The answer to the question posed in the title is \textbf{yes} (NLI models can validate these examples), with two caveats. First, it is helpful to create QA-specific data for the NLI model. Second, the information that is sufficient for a question to be fully answered may not align with annotations in the QA dataset. We encourage further explorations of the interplay between these tasks and careful analysis of the predictions of QA models.

\section*{Acknowledgments}

This work was partially supported by NSF Grant IIS-1814522. We would like to thank Kaj Bostrom, Yasumasa Onoe, and the anonymous reviewers for their helpful comments.

This material is also based on research that is in part supported by the Air Force Research Laboratory (AFRL), DARPA, for the KAIROS program under agreement number FA8750-19-2-1003. The U.S. Government is authorized to reproduce and distribute reprints for Governmental purposes notwithstanding any copyright notation thereon. The views and conclusions contained herein are those of the authors and should not be interpreted as necessarily representing the official policies or endorsements, either expressed or implied, of the Air Force Research Laboratory (AFRL), DARPA, or the U.S. Government.

\bibliography{anthology,custom}
\bibliographystyle{acl_natbib}

\appendix

\section{Statistics of the Converted Datasets}
\label{section:appendix:dataset_statistics}
The statistics of the datasets after processing through our pipeline is shown in Table~\ref{tab:word_overlap}. Both the premise length and the hypothesis length are quite similar except for the premise length of TriviaQA, despite their original context length differs greatly~\cite{fisch-etal-2019-mrqa}.

\section{Model Details}
\label{section:appendix:model_details}

\subsection{Answer Generator}
We train our \texttt{Bert-joint} on the full NQ training set for 1 epoch. We initialize the model with \texttt{bert-large-uncased-wwm}.\footnote{https://github.com/google-research/bert} The batch size is set to 8, window size is set to 512, and the optimizer we use is Adam~\cite{kingma2015adam} with initial learning rate setting to 3e-5.

\subsection{Question Converter}
Each instance of the input is constructed as $[\text{CLS}] q [\text{S}] a [\text{S}]$, where [CLS] and [S] are the classification and separator tokens of the T5 model respectively. The output is the target sentence $d$.

The model is trained using the seq2seq framework of Huggingface~\cite{wolf-etal-2020-transformers}. The max source sequence length is set to 256 and the target sequence length is set to 512. Batch size is set to 12 and we use Deepspeed for memory optimization~\cite{rasley2020deepspeed}. We train the model with 86k question-answer pairs for 1 epoch with Adam optimizer and an initial learning rate set to 3e-5. 95\% of question answer pairs come from SQuAD and the remaining 5\% come from four other question answering datasets~\cite{demszky2018transforming}.

\subsection{Decontextualizer}
Each instance of the input is constructed as follows:

$\text{[CLS]} \text{T} \text{[S]} x_1, ..., x_{t-1} \text{[S]} x_t \text{[S]} x_{t+1}, ..., x_n \text{[S]}$
where [CLS] and [S] are the classification and separator tokens of the T5 model respectively. T denotes the context title which could be empty. $x_i$ denotes the $i$th sentence in the context and $x_t$ is the target sentence to decontextualize.

The model is trained using the seq2seq framework of Huggingface~\cite{wolf-etal-2020-transformers}. The max sequence length for both source and target is set to 512. Batch size is set to 4 and we use Deepspeed for memory optimization~\cite{rasley2020deepspeed}. We train the model with 11k question-answer pairs~\cite{choi2021decontextualization} for 5 epoch with Adam optimizer and an initial learning rate set to 3e-5. 

\subsection{NQ-NLI}
The generated NQ-NLI training and development set contain 191k and 4,855 (premise, hypothesis) pairs from NQ respectively. We initialize the model with \texttt{roberta-large}~\cite{liu2019roberta} and train the model for 5 epochs. Batch size is set to 16, with Adam as the optimizer and initial learning rate set to 2e-6.

\begin{table}[t]
\small
\centering
\begin{tabular}{ c | c c c }
\toprule
 & Prem Len &  Hyp Len & Word Overlap \\
\midrule
NQ & 20.0 & 8.0 & 0.22  \\
\midrule
TriviaQA & 15.9 & 9.0 & 0.16 \\
BioASQ & 20.6 &	8.0 & 0.14 \\
\midrule
SQuAD 2.0 & 19.1 & 8.2 & 0.23 \\
SQuAD-adv & 19.0 & 8.2 & 0.26 \\

\bottomrule
\end{tabular}
\caption{Statistics of the development set for each dataset listed above. Here, ``Prem len'' and ``Hyp len'' denote the average number of words with stop words removed in the premise and hypothesis respectively; ``Word Overlap'' denotes the Jaccard similarity between the premise and the hypothesis.}
\label{tab:word_overlap}
\end{table}

\section{Performance Breakdown on All Datasets}
\label{section:appendix:performance-breakdown}

\begin{figure*}[t]
\centering
\includegraphics[width=1.0\textwidth]{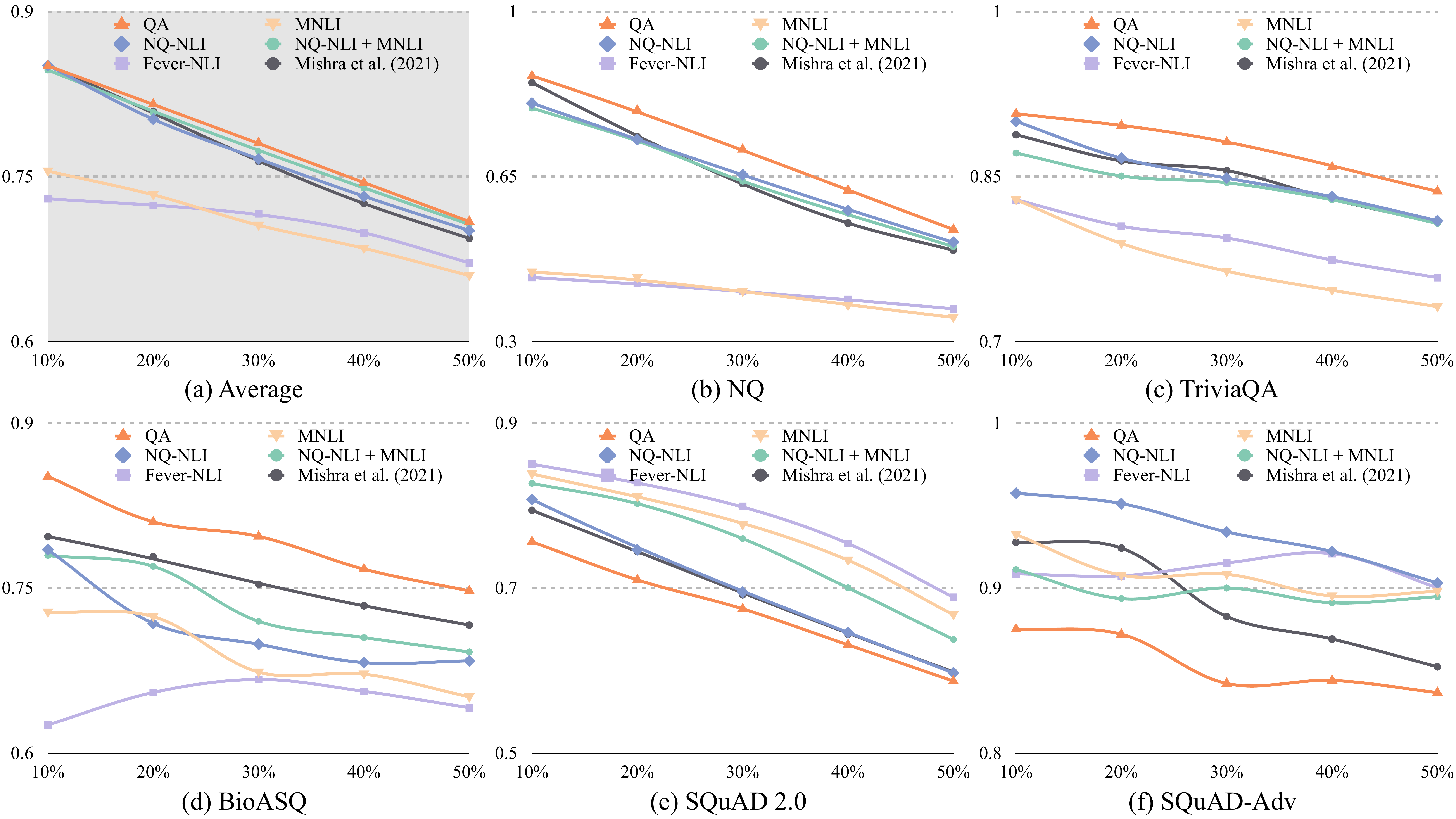}
\caption{Calibration performance of the NQ-NLI models on five QA datasets we used in the paper. The training using NQ-NLI helps close the gap between the QA and the NLI models. The x-axis denotes coverage and the y-axis denotes the F1 score.}
    \label{fig:appendix:NQ-NLI-fine-tuning}
\end{figure*}

\begin{figure*}[t]
\centering
\includegraphics[width=1.0\textwidth]{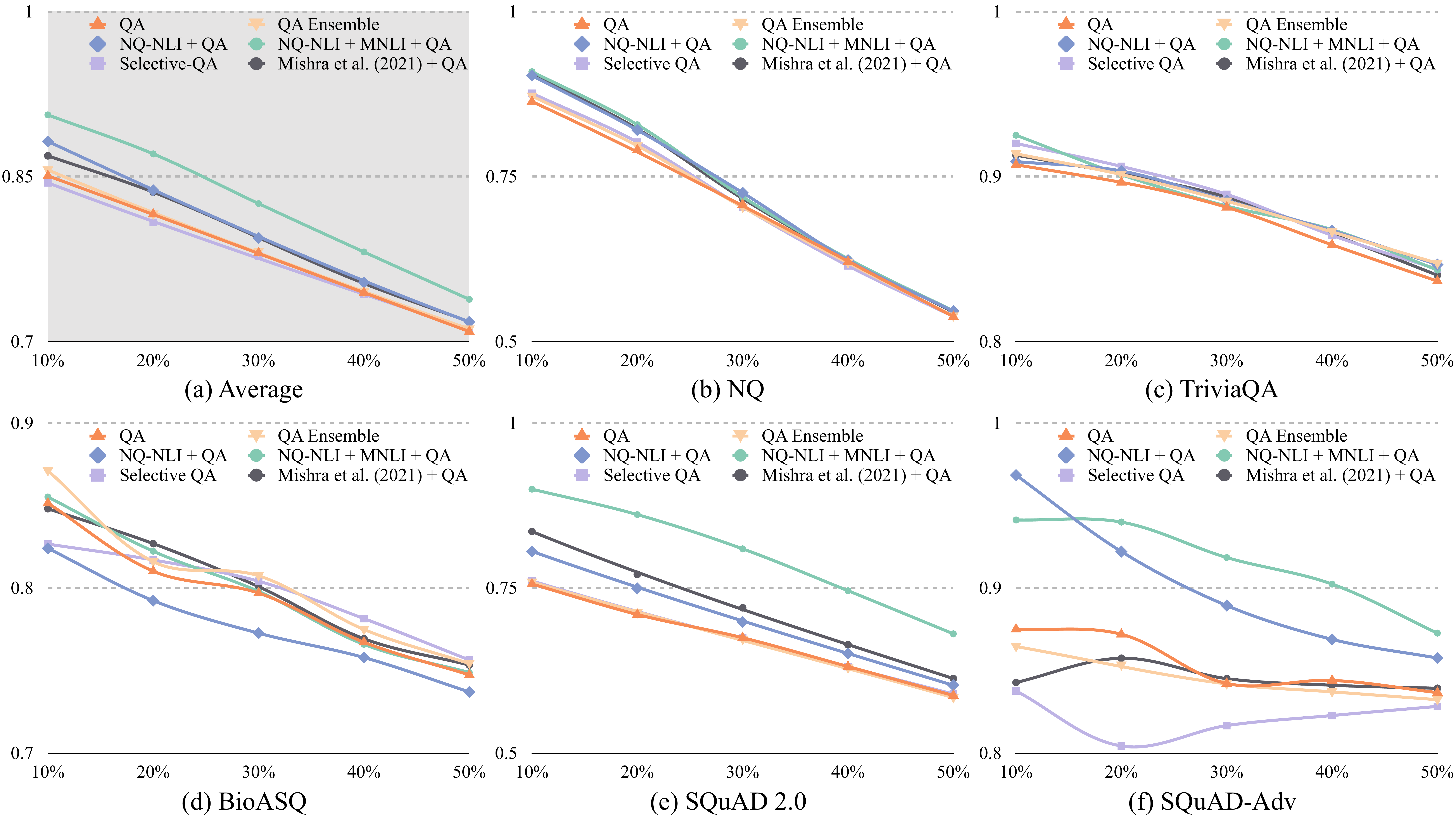}
\caption{Calibration performance of the NQ-NLI models combined with the QA model on five QA datasets we used in the paper. The combined \textbf{NQ-NLI+MNLI+QA} model largely outperforms the QA model on all datasets. The x-axis denotes coverage and the y-axis denotes the F1 score.}
    \label{fig:appendix:NQ-NLI-fine-tuning-plus-qa}
\end{figure*}

Figures~\ref{fig:appendix:NQ-NLI-fine-tuning} and \ref{fig:appendix:NQ-NLI-fine-tuning-plus-qa} show full results for Figures~\ref{fig:exp_qa_nli_fine_tuning_avg} and \ref{fig:exp_qa_nli_fine_tuning_plus_qa_avg}, respectively.

\end{document}